\documentclass{article}
\ifdefined\pdfsuppressptexinfo\pdfsuppressptexinfo=-1\fi
\ifdefined\pdftrailerid\pdftrailerid{}\fi

\usepackage[T1]{fontenc}
\usepackage{microtype}
\usepackage{graphicx}
\usepackage{booktabs}
\usepackage{array}
\usepackage{hyperref}
\usepackage{xurl}
\usepackage[preprint]{icml2027}
\makeatletter
\renewcommand{\ICML@preprint}{\textit{Preprint. November 2026.}}
\makeatother

\usepackage{amsmath}
\usepackage{amssymb}
\usepackage[capitalize,noabbrev]{cleveref}

\newcommand{\auDistilbertSstTwoOrtCpuIntEightDynamic}{0.98}
\newcommand{\auDistilbertSstTwoOrtCpuIntEightDynamichi}{0.99}
\newcommand{\auDistilbertSstTwoOrtCpuIntEightDynamiclo}{0.96}

\newcommand{\auHiDeitTiny}{0.96}

\newcommand{\auHiEfficientnetBZero}{0.85}
\newcommand{\auHiMobilenetVThreeLarge}{0.92}
\newcommand{\auHiResnetFiveZero}{0.98}
\newcommand{\auLoDeitTiny}{0.91}

\newcommand{\auLoEfficientnetBZero}{0.61}
\newcommand{\auLoMobilenetVThreeLarge}{0.67}
\newcommand{\auLoResnetFiveZero}{0.89}
\newcommand{\auMaxDeitTiny}{0.96}

\newcommand{\auMaxEfficientnetBZero}{0.84}
\newcommand{\auMaxMobilenetVThreeLarge}{0.92}
\newcommand{\auMaxResnetFiveZero}{0.97}
\newcommand{\auMinDeitTiny}{0.92}

\newcommand{\auMinEfficientnetBZero}{0.63}
\newcommand{\auMinMobilenetVThreeLarge}{0.68}
\newcommand{\auMinResnetFiveZero}{0.90}

\newcommand{\byNewMax}{0.5}

\newcommand{\byRepDeitTiny}{95.7}
\newcommand{\byRepEfficientnetBZero}{99.5}
\newcommand{\byRepMobilenetVThreeLarge}{99.0}
\newcommand{\byRepResnetFiveZero}{96.9}

\newcommand{\cpDistilbertSstTwoZero}{0.15}

\newcommand{\cpMobilenetVThreeLargeTrtTfThreeTwo}{0.20}

\newcommand{\cpResnetFiveZeroZero}{0.04}

\newcommand{\daDeitTinyLtPtTwoEIntEightCZero}{\ensuremath{-}0.06}

\newcommand{\daDistilbertSstTwoOrtCpuIntEightQdqCZero}{\ensuremath{-}38.49}

\newcommand{\daMaxEfficientnetBZero}{\ensuremath{-}9.33}
\newcommand{\daMaxMobilenetVThreeLarge}{\ensuremath{-}2.57}

\newcommand{\daMinEfficientnetBZero}{\ensuremath{-}84.25}
\newcommand{\daMinMobilenetVThreeLarge}{\ensuremath{-}55.35}

\newcommand{\daResnetFiveZeroOrtCpuIntEightQdqCZero}{\ensuremath{-}0.15}

\newcommand{\decOneDeitTinyLtPtTwoEIntEightCZero}{84}
\newcommand{\decOneDeitTinyLtPtTwoEIntEightCZerohi}{88}
\newcommand{\decOneDeitTinyLtPtTwoEIntEightCZerolo}{80}

\newcommand{\flDeitTinyLtPtTwoEIntEightCZero}{3.65}
\newcommand{\flDeitTinyLtPtTwoEIntEightCZerohi}{4.07}
\newcommand{\flDeitTinyLtPtTwoEIntEightCZerolo}{3.28}

\newcommand{\flDistilbertSstTwoEtXnnIntEightDynamic}{0.26}
\newcommand{\flDistilbertSstTwoEtXnnIntEightDynamichi}{0.52}
\newcommand{\flDistilbertSstTwoEtXnnIntEightDynamiclo}{0.05}

\newcommand{\flDistilbertSstTwoLtLqWiEightAfpThreeTwo}{0.67}
\newcommand{\flDistilbertSstTwoLtLqWiEightAfpThreeTwohi}{1.03}
\newcommand{\flDistilbertSstTwoLtLqWiEightAfpThreeTwolo}{0.31}

\newcommand{\flEfficientnetBZeroEtXnnIntEightCZero}{16.97}

\newcommand{\flEfficientnetBZeroTrtIntEightQdqCZero}{98.13}

\newcommand{\flMobilenetVThreeLargeTrtFpOneSix}{0.33}
\newcommand{\flMobilenetVThreeLargeTrtFpOneSixhi}{0.47}
\newcommand{\flMobilenetVThreeLargeTrtFpOneSixlo}{0.22}

\newcommand{\flMobilenetVThreeLargeTrtTfThreeTwo}{0.11}

\newcommand{\flResnetFiveZeroEtXnnIntEightCZero}{2.48}
\newcommand{\flResnetFiveZeroEtXnnIntEightCZerohi}{2.84}
\newcommand{\flResnetFiveZeroEtXnnIntEightCZerolo}{2.16}

\newcommand{\flResnetFiveZeroLtPtTwoEIntEightCZero}{2.42}
\newcommand{\flResnetFiveZeroLtPtTwoEIntEightCZerohi}{2.78}
\newcommand{\flResnetFiveZeroLtPtTwoEIntEightCZerolo}{2.10}

\newcommand{\flResnetFiveZeroOrtCpuIntEightQdqCZero}{4.07}
\newcommand{\flResnetFiveZeroOrtCpuIntEightQdqCZerohi}{4.53}
\newcommand{\flResnetFiveZeroOrtCpuIntEightQdqCZerolo}{3.65}

\newcommand{\kfDeitTinyLtPtTwoEIntEight}{4.16}
\newcommand{\kfDeitTinyLtPtTwoEIntEighthi}{4.63}
\newcommand{\kfDeitTinyLtPtTwoEIntEightlo}{3.74}

\newcommand{\kfResnetFiveZeroOrtCpuIntEightQdq}{3.44}
\newcommand{\kfResnetFiveZeroOrtCpuIntEightQdqhi}{3.88}
\newcommand{\kfResnetFiveZeroOrtCpuIntEightQdqlo}{3.06}

\newcommand{\liftMinEfficientnetBZero}{1.0}
\newcommand{\liftMinEfficientnetBZerohi}{1.0}
\newcommand{\liftMinEfficientnetBZerolo}{1.0}
\newcommand{\liftMinMobilenetVThreeLarge}{1.2}
\newcommand{\liftMinMobilenetVThreeLargehi}{1.3}
\newcommand{\liftMinMobilenetVThreeLargelo}{1.2}
\newcommand{\liftMinResnetFiveZero}{6.8}
\newcommand{\liftMinResnetFiveZerohi}{7.6}
\newcommand{\liftMinResnetFiveZerolo}{6.0}

\newcommand{\liftResnetFiveZeroEtXnnIntEightCZeroLtPtTwoEIntEightCZero}{31.8}
\newcommand{\liftResnetFiveZeroEtXnnIntEightCZeroLtPtTwoEIntEightCZerohi}{36.0}
\newcommand{\liftResnetFiveZeroEtXnnIntEightCZeroLtPtTwoEIntEightCZerolo}{28.2}

\newcommand{\maxAbsFpThreeTwo}{4.5\times 10^{-3}}
\newcommand{\nEqPrimary}{4}
\newcommand{\nEqPrimaryAboveOne}{4}
\newcommand{\nEvalImages}{7{,}854}
\newcommand{\nEvalText}{1{,}938}
\newcommand{\nFpThreeTwoConv}{34}

\newcommand{\nModels}{5}
\newcommand{\nPosText}{6}
\newcommand{\nPosVision}{40}
\newcommand{\nPrimaryVision}{14}

\newcommand{\nPtsVision}{75}
\newcommand{\nRecallText}{6}
\newcommand{\nRecallVision}{40}
\newcommand{\nStacksRun}{116}
\newcommand{\nZeroText}{8}
\newcommand{\nZeroVision}{27}

\newcommand{\nbResnetFiveZeroOrtCpuIntEightQdqCZero}{102}

\newcommand{\nfMobilenetVThreeLargeTrtTfThreeTwo}{9}

\newcommand{\nwResnetFiveZeroOrtCpuIntEightQdqCZero}{114}

\newcommand{\plResnetFiveZeroLtPtTwoEIntEightCZero}{51}

\newcommand{\prResnetFiveZeroLtPtTwoEIntEightCZero}{10}

\newcommand{\recallRndText}{0.67}
\newcommand{\recallRndTexthi}{0.96}
\newcommand{\recallRndTextlo}{0.22}
\newcommand{\recallRndVision}{0.95}
\newcommand{\recallRndVisionhi}{0.99}
\newcommand{\recallRndVisionlo}{0.83}

\newcommand{\recallTexthi}{1.00}
\newcommand{\recallTextlo}{0.54}

\newcommand{\recallVisionhi}{1.00}
\newcommand{\recallVisionlo}{0.91}
\newcommand{\redrawRepDeitTiny}{46}
\newcommand{\redrawRepEfficientnetBZero}{10}
\newcommand{\redrawRepMobilenetVThreeLarge}{31}
\newcommand{\redrawRepResnetFiveZero}{41}
\newcommand{\rhoPText}{0.999}
\newcommand{\rhoPVision}{0.988}
\newcommand{\rhoRndText}{0.838}
\newcommand{\rhoRndVision}{0.958}

\newcommand{\tyBestDeitTiny}{Add}

\newcommand{\tyConvEfficientnetBZero}{49}
\newcommand{\tyConvMobilenetVThreeLarge}{49}
\newcommand{\tyConvResnetFiveZero}{50}
\newcommand{\tyMaxAll}{61}
\newcommand{\tyMaxDeitTiny}{55}
\newcommand{\tyMaxEfficientnetBZero}{49}
\newcommand{\tyMaxMobilenetVThreeLarge}{61}
\newcommand{\tyMaxResnetFiveZero}{61}

\newcommand{\exRef}{golden retriever}
\newcommand{\exAlt}{cocker spaniel}
\newcommand{\exPrefRef}{0.14}
\newcommand{\exPrefAlt}{0.12}
\newcommand{\exPaltRef}{0.12}
\newcommand{\exPaltAlt}{0.16}
\newcommand{\exN}{74}

\icmltitlerunning{One Checkpoint, Four Converters}

\hypersetup{pdftitle={One Checkpoint, Four Converters: Per Input Disagreement After On Device Conversion and a Probe Set That Catches It}, pdfauthor={Sharven Rane}, pdfsubject={Preprint}, pdfkeywords={model conversion, ExecuTorch, LiteRT, TensorRT, ONNX Runtime, quantization, conformance testing}, pdfcreator={LaTeX}}
\begin{document}
\twocolumn[
  \icmltitle{One Checkpoint, Four Converters: Per Input Disagreement\\
    After On Device Conversion and a Probe Set That Catches It}
  \begin{icmlauthorlist}\icmlauthor{Sharven Rane}{ind}\end{icmlauthorlist}
  \icmlaffiliation{ind}{Independent Researcher}
  \icmlcorrespondingauthor{Sharven Rane}{sharvenrane7@gmail.com}
  \icmlkeywords{model conversion, ExecuTorch, LiteRT, TensorRT, ONNX Runtime, quantization, conformance testing}
  \vskip 0.3in
]
\printAffiliationsAndNotice{}

\begin{abstract}
A model converted for on device or accelerated inference ships once its top 1 accuracy matches the source model. Accuracy can match while individual decisions change, and the converter, its INT8 recipe and its calibration draw each move them. We convert \nModels{} public checkpoints (four ImageNet classifiers and a sentiment model) with the documented FP32 and INT8 recipes of ExecuTorch, LiteRT, ONNX Runtime and TensorRT and compare every top 1 decision with the PyTorch source on \nEvalImages{} images and \nEvalText{} sentences. All \nFpThreeTwoConv{} strict FP32 conversions reproduce every decision, yet each of the \nEqPrimary{} vision INT8 recipes that keeps accuracy equivalent flips more than 1\% of decisions: ONNX Runtime static INT8 flips \flResnetFiveZeroOrtCpuIntEightQdqCZero\% (95\% CI [\flResnetFiveZeroOrtCpuIntEightQdqCZerolo, \flResnetFiveZeroOrtCpuIntEightQdqCZerohi]) of ResNet 50 decisions while accuracy moves by \daResnetFiveZeroOrtCpuIntEightQdqCZero{} points, and two builds of that recipe from different calibration draws disagree on \kfResnetFiveZeroOrtCpuIntEightQdq\% [\kfResnetFiveZeroOrtCpuIntEightQdqlo, \kfResnetFiveZeroOrtCpuIntEightQdqhi]. Where accuracy is kept, the flipped inputs are the ones the source model barely decides (AUROC of the source margin \auMinResnetFiveZero{} to \auMaxResnetFiveZero{} on ResNet 50, every 95\% CI inside [\auLoResnetFiveZero, \auHiResnetFiveZero]), but our preregistered claim that this holds for every recipe fails on the two mobile networks, where INT8 breaks accuracy outright. A probe set of the 200 training inputs with the smallest source margin, chosen before any conversion, flags all \nPosVision{} vision stacks (recall 95\% CI [\recallVisionlo, \recallVisionhi]) and all \nPosText{} text stacks that flip at least 0.5\% of decisions and none of the exact FP32 conversions, so conversion checks should compare decisions on such a probe set, not accuracy.
\end{abstract}

\begin{figure*}[t]
  \centering
  \includegraphics[width=\textwidth]{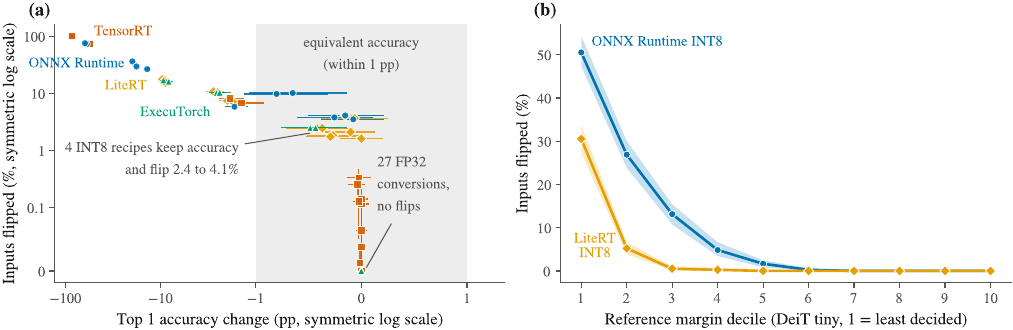}
  \caption{INT8 conversions that keep top 1 accuracy within 1 point still flip \flResnetFiveZeroLtPtTwoEIntEightCZero{} to \flResnetFiveZeroOrtCpuIntEightQdqCZero\% of decisions, and the
  flips sit where the source model barely decides. (a) Every converted vision stack of four models: top 1
  accuracy change against the share of inputs whose decision differs from PyTorch FP32; bars are 95\%
  bootstrap intervals over \nEvalImages{} images; the shaded band is the preregistered equivalence margin.
  Strict FP32 conversions sit at the origin. (b) DeiT tiny, share of inputs flipped by the two INT8 recipes
  that run on it, by decile of the source margin (top 1 minus top 2 probability); bands are 95\% bootstrap intervals.}
  \label{fig:one}
\end{figure*}

\section{Introduction}\label{sec:intro}

A trained model rarely runs where it was trained. Phones run it through LiteRT, Core ML or ExecuTorch,
servers through ONNX Runtime or TensorRT, and the step that gets it there is a converter followed, more
often than not, by INT8 post training quantization. Teams accept a converted model when its top 1 accuracy
on a held out set matches the source model within a tolerance. That check is cheap and it is what the
converters' own tutorials show.

Accuracy can match while the decisions underneath it change. Take one ResNet 50 image from our evaluation
set: PyTorch calls it a \exRef{} with probability \exPrefRef{} against \exPrefAlt{} for \exAlt; the same
checkpoint converted with LiteRT's documented static INT8 recipe calls it a \exAlt{} (\exPaltAlt{} against
\exPaltRef). The image is not exotic: it is the median of the \exN{} such swaps that recipe makes on this
model. Every one of them is a user who sees a different answer after the update, and accuracy hides them
because swaps in both directions cancel: the ONNX Runtime INT8 build of ResNet 50 fixes
\nbResnetFiveZeroOrtCpuIntEightQdqCZero{} source errors and introduces \nwResnetFiveZeroOrtCpuIntEightQdqCZero.

We measure this per input disagreement for one fixed checkpoint across four converters. The measurement
is easy to get wrong in both directions. A harness that is not bitwise reproducible invents flips; on the
conversion side, a TF32 default silently changes FP32 numerics and an INT8 build depends on which 256
calibration inputs were drawn. We therefore show first that the instrument reads zero where it should
(\cref{sec:kill}), report every rate with an interval next to a calibration redraw floor, and fixed the
claims, thresholds and stacks in a preregistration before the main runs.

Strict FP32 conversions reproduce all decisions. The INT8 recipes split into two groups: on MobileNetV3
and EfficientNet they lose accuracy outright, which any accuracy check catches; on ResNet 50 and DeiT they
keep accuracy equivalent and flip \flResnetFiveZeroLtPtTwoEIntEightCZero\% to
\flResnetFiveZeroOrtCpuIntEightQdqCZero\% of decisions, which no accuracy check catches. A probe set picked
from the source model alone catches both.

\textbf{Contributions.}
\begin{itemize}
\item \textbf{Every vision INT8 recipe that keeps accuracy flips more than 1\% of decisions}:
\nEqPrimaryAboveOne{} of \nEqPrimary{} equivalent primary recipes, up to \flResnetFiveZeroOrtCpuIntEightQdqCZero\%
[\flResnetFiveZeroOrtCpuIntEightQdqCZerolo, \flResnetFiveZeroOrtCpuIntEightQdqCZerohi] at an accuracy change
of \daResnetFiveZeroOrtCpuIntEightQdqCZero{} points, while all \nFpThreeTwoConv{} strict FP32 conversions
flip none (\cref{fig:one}a, \cref{tab:int8}, \cref{sec:c1}).
\item \textbf{The source margin locates the flips only where accuracy is kept}: the preregistered claim that it
predicts flips for every recipe (AUROC at least 0.85) and that flip sets overlap across converters
(lift at least 3) is refuted on MobileNetV3 and EfficientNet and holds on ResNet 50 and DeiT
(\cref{fig:one}b, \cref{fig:overlap}, \cref{sec:c2}).
\item \textbf{200 low margin probes chosen before conversion flag every nonconformant stack}: recall
\nRecallVision{} of \nPosVision{} vision stacks (95\% CI [\recallVisionlo, \recallVisionhi]) and \nRecallText{} of \nPosText{}
text stacks ([\recallTextlo, \recallTexthi]), no false alarm on
\nZeroVision{} plus \nZeroText{} exact conversions, and probe flip rates rank stacks like the full
evaluation set (Spearman \rhoPVision{} for vision) (\cref{fig:probe}, \cref{sec:c3}).
\end{itemize}
We also report five converter defects found while running the documented recipes, two of which pass
export and fail only when the runtime loads the model (\cref{tab:defects}).

\section{Setup}\label{sec:setup}

\paragraph{Models and inputs.} We convert \nModels{} public checkpoints: MobileNetV3 Large, EfficientNet B0
and ResNet 50 from torchvision, DeiT tiny, and DistilBERT fine tuned on SST-2. Vision models are scored on
the validation splits of Imagenette and Imagewoof (\nEvalImages{} images of ten easy and ten fine grained
ImageNet classes) with their full 1000 way output; DistilBERT on SST-2 validation and Rotten Tomatoes test
(\nEvalText{} sentences). Preprocessing is computed once in numpy, so every runtime receives the same
float32 tensor bit for bit. Calibration uses two disjoint draws of 256 training inputs; a further 2{,}000
training inputs form the probe pool of \cref{sec:c3}. All three sets are disjoint from evaluation.

\paragraph{Decisions and flips.} The decision is the top 1 class. The reference stack is PyTorch 2.11 eager
on CPU in FP32 at batch size 1. A flip is an input whose decision under a converted stack differs from the
reference decision on the same input; the flip rate is flips over inputs. Inputs are independent, so
intervals are a bootstrap over inputs (2000 resamples) and zero counts get the exact one sided 95\%
Clopper Pearson bound. Accuracy is called equivalent when the paired 95\% interval of its change lies
within $\pm 1$ percentage point (pp).

\paragraph{Converters and recipes.} Each converter runs its documented path and, for INT8, its documented
post training recipe: ONNX Runtime 1.29 static QDQ (per channel, MinMax) and dynamic quantization;
TensorRT 10.16 FP32, TF32, FP16 and explicit INT8 from a symmetric QDQ graph; ExecuTorch 1.2 with the
XNNPACK delegate in FP32 and PT2E INT8, plus its portable kernels; LiteRT through litert\_torch in FP32,
PT2E static and dynamic INT8, and the LiteRT Quantizer recipe with int8 weights and float activations. The
primary INT8 recipe of each converter is fixed in advance (static per channel for vision; dynamic for
text, where it is the documented transformer recipe). Core ML models convert with coremltools but cannot
run off macOS, so Core ML enters only as conversion success. ExecuTorch and LiteRT run their x86 CPU
kernels, the kernel families phones use; this is a CPU proxy, not a phone measurement. In total we
evaluate \nStacksRun{} stacks; stacks that fail to build or load are listed in \cref{tab:defects} and
excluded from rates. ExecuTorch portable kernels need hours per pass, so they ran on four of the five models (not ResNet 50).

\paragraph{Claims and decision rules.} The preregistration fixes three claims. C1: among vision INT8
primary recipes with equivalent accuracy, at least half have a flip rate whose lower bound exceeds 1\% and
5 times the FP32 noise floor. C2: (a) the AUROC of the negative source margin for predicting a flip is at
least 0.85 for every vision INT8 primary recipe; (b) every pair of primary recipes on a model shares
flips at least 3 times more often than independence predicts (lift). C3: a probe set of the 200 lowest
margin inputs from the probe pool ranks stacks like the evaluation set (Spearman at least 0.9), and the
rule ``flag a stack if any probe flips'' catches every stack with an evaluation flip rate of at least 0.5\%
with at most one false alarm among the exact FP32 conversions, while 200 random probes catch fewer.

\section{The instrument works}\label{sec:kill}

\cref{tab:fp32} shows the controls before any claim. A rerun of the reference gives bitwise identical
logits on all five models (K1). Batch size 32, one thread, PyTorch 2.14 and CUDA FP32 without TF32 flip
no decision on any model, so the FP32 noise floor is zero (K2). All \nFpThreeTwoConv{} strict FP32
conversions also flip no decision (K3); with \nEvalImages{} images the one sided bound is
\cpResnetFiveZeroZero\% per vision stack and \cpDistilbertSstTwoZero\% for text. The largest logit
difference among them is $\maxAbsFpThreeTwo$, too small to cross a decision on this data.

One preregistered FP32 stack does flip. TensorRT's default build allows TF32 matrix units, and it flips
\nfMobilenetVThreeLargeTrtTfThreeTwo{} MobileNetV3 decisions (\flMobilenetVThreeLargeTrtTfThreeTwo\%, bound
\cpMobilenetVThreeLargeTrtTfThreeTwo\%): above the 0.1\% line, below the 0.3\% bound, and the documented
behaviour of that flag rather than a defect. A team that converts ``in FP32'' with defaults has already
flipped decisions. TensorRT FP16 flips at most \flMobilenetVThreeLargeTrtFpOneSix\% [\flMobilenetVThreeLargeTrtFpOneSixlo, \flMobilenetVThreeLargeTrtFpOneSixhi] (MobileNetV3) and none
on DistilBERT.

The last floor is the calibration draw (K4). Two builds of the same INT8 recipe from two disjoint
256 input draws disagree with each other on \kfResnetFiveZeroOrtCpuIntEightQdq\% [\kfResnetFiveZeroOrtCpuIntEightQdqlo, \kfResnetFiveZeroOrtCpuIntEightQdqhi] of ResNet 50 inputs (ONNX
Runtime) and \kfDeitTinyLtPtTwoEIntEight\% [\kfDeitTinyLtPtTwoEIntEightlo, \kfDeitTinyLtPtTwoEIntEighthi] of DeiT inputs (LiteRT), about as often as either disagrees with
the source (last column of \cref{tab:int8}). An INT8 flip rate is therefore a property of a recipe and a
calibration sample, not of the converter alone.

\begin{table}[t]
\caption{Every strict FP32 conversion reproduces every PyTorch decision. Cells: flips
against PyTorch CPU FP32 (7{,}854 images per vision model, 1{,}938 sentences);
last row: exact one sided 95\% Clopper Pearson bound at 0 flips. n/r: not run (B4, B5, run time).
ORT ONNX Runtime, TRT TensorRT, ET ExecuTorch.}
\label{tab:fp32}
\centering\footnotesize\setlength{\tabcolsep}{3.5pt}
\begin{tabular}{lrrrrr}
\toprule
Stack & MNv3 & EffB0 & R50 & DeiT & DBERT \\
\midrule
\multicolumn{6}{l}{\emph{Controls (K1, K2)}} \\
PyTorch rerun (K1) & 0 & 0 & 0 & 0 & 0 \\
PyTorch batch 32 & 0 & 0 & 0 & 0 & 0 \\
PyTorch 1 thread & 0 & 0 & 0 & 0 & 0 \\
PyTorch 2.14 & 0 & 0 & 0 & 0 & 0 \\
PyTorch CUDA FP32 & 0 & 0 & 0 & 0 & 0 \\
\midrule
\multicolumn{6}{l}{\emph{FP32 conversions (K3)}} \\
ORT CPU & 0 & 0 & 0 & 0 & 0 \\
ORT CPU, no opt. & 0 & 0 & 0 & 0 & 0 \\
ORT CUDA & 0 & 0 & 0 & 0 & 0 \\
TRT FP32 & 0 & 0 & 0 & 0 & 0 \\
ET XNNPACK & 0 & 0 & 0 & 0 & 0 \\
ET portable & 0 & 0 & n/r & 0 & 0 \\
LiteRT & 0 & 0 & 0 & 0 & 0 \\
\midrule
TRT default (TF32) & 9 & 3 & 1 & 5 & 0 \\
TRT FP16 & 26 & 20 & 10 & 11 & 0 \\
\midrule
Bound at 0 flips (\%) & 0.04 & 0.04 & 0.04 & 0.04 & 0.15 \\
\bottomrule
\end{tabular}
\end{table}

\section{Results}\label{sec:results}

\subsection{INT8 conversions flip decisions that accuracy does not show}\label{sec:c1}

\cref{tab:int8} and \cref{fig:one}a give every INT8 primary recipe. On MobileNetV3 and EfficientNet no
INT8 recipe is equivalent: accuracy changes by \daMaxMobilenetVThreeLarge{} to \daMinMobilenetVThreeLarge{}
points on MobileNetV3 and \daMaxEfficientnetBZero{} to \daMinEfficientnetBZero{} on EfficientNet. These
conversions are broken in a way an accuracy check sees, and C1 conditions on equivalence, so they neither
support nor refute it. The spread across converters on the same checkpoint is itself large: the same
EfficientNet weights keep \flEfficientnetBZeroEtXnnIntEightCZero\% flips with ExecuTorch and
\flEfficientnetBZeroTrtIntEightQdqCZero\% with TensorRT.

On ResNet 50 and DeiT, \nEqPrimary{} of the \nPrimaryVision{} vision primary recipes are equivalent, and all
\nEqPrimaryAboveOne{} have a flip rate lower bound above 1\% with a noise floor of zero, so C1 holds. ONNX
Runtime flips \flResnetFiveZeroOrtCpuIntEightQdqCZero\% [\flResnetFiveZeroOrtCpuIntEightQdqCZerolo,
\flResnetFiveZeroOrtCpuIntEightQdqCZerohi] of ResNet 50 decisions at \daResnetFiveZeroOrtCpuIntEightQdqCZero{}
points, ExecuTorch \flResnetFiveZeroEtXnnIntEightCZero\% [\flResnetFiveZeroEtXnnIntEightCZerolo, \flResnetFiveZeroEtXnnIntEightCZerohi] and LiteRT
\flResnetFiveZeroLtPtTwoEIntEightCZero\% [\flResnetFiveZeroLtPtTwoEIntEightCZerolo, \flResnetFiveZeroLtPtTwoEIntEightCZerohi];
LiteRT flips \flDeitTinyLtPtTwoEIntEightCZero\% [\flDeitTinyLtPtTwoEIntEightCZerolo,
\flDeitTinyLtPtTwoEIntEightCZerohi] of DeiT decisions at \daDeitTinyLtPtTwoEIntEightCZero{} points. On text,
the dynamic recipes of ExecuTorch and LiteRT stay equivalent and flip \flDistilbertSstTwoEtXnnIntEightDynamic\%
[\flDistilbertSstTwoEtXnnIntEightDynamiclo, \flDistilbertSstTwoEtXnnIntEightDynamichi] and \flDistilbertSstTwoLtLqWiEightAfpThreeTwo\%
[\flDistilbertSstTwoLtLqWiEightAfpThreeTwolo, \flDistilbertSstTwoLtLqWiEightAfpThreeTwohi]; ONNX Runtime static QDQ breaks DistilBERT
(\daDistilbertSstTwoOrtCpuIntEightQdqCZero{} points).

\begin{table*}[t]
\caption{Vision INT8 recipes either break accuracy or keep it while flipping 2.4 to 4.1\% of decisions; builds from two disjoint 256 input calibration draws disagree about as often (K4, last column).
95\% bootstrap CIs over 7{,}854 images or 1{,}938 sentences. EQ: accuracy change CI inside $\pm 1$ pp. Bold: EQ and flip rate lower bound
above 1\% (C1). n/a: no calibration; missing rows: B4, B5.}
\label{tab:int8}
\centering\footnotesize\setlength{\tabcolsep}{5pt}
\begin{tabular}{llr@{\,}lr@{\,}lcr@{\,}l}
\toprule
Model & INT8 recipe & \multicolumn{2}{c}{Flip rate (\%)} & \multicolumn{2}{c}{Accuracy change (pp)} & EQ & \multicolumn{2}{c}{Calibration redraw (\%)} \\
 & & & [95\% CI] & & [95\% CI] & & & [95\% CI] \\
\midrule
MobileNetV3 Large & ORT static QDQ & 26.48 & [25.54, 27.46] & $-$13.81 & [$-$14.77, $-$12.92] & no & 29.25 & [28.25, 30.25] \\
 & TRT explicit INT8 & 73.07 & [72.10, 74.05] & $-$55.35 & [$-$56.46, $-$54.20] & no & 76.17 & [75.20, 77.09] \\
 & ET XNNPACK PT2E & 10.19 & [9.50, 10.90] & $-$2.57 & [$-$3.12, $-$2.02] & no & 9.33 & [8.70, 10.00] \\
 & LiteRT PT2E static & 10.68 & [9.99, 11.40] & $-$2.76 & [$-$3.34, $-$2.16] & no & 9.63 & [8.95, 10.29] \\
\midrule
EfficientNet B0 & ORT static QDQ & 72.93 & [72.00, 73.95] & $-$59.69 & [$-$60.85, $-$58.65] & no & 67.90 & [66.86, 68.89] \\
 & TRT explicit INT8 & 98.13 & [97.84, 98.43] & $-$84.25 & [$-$85.04, $-$83.44] & no & 93.54 & [93.01, 94.08] \\
 & ET XNNPACK PT2E & 16.97 & [16.14, 17.85] & $-$9.33 & [$-$10.12, $-$8.54] & no & 12.55 & [11.84, 13.38] \\
 & LiteRT PT2E static & 17.44 & [16.67, 18.30] & $-$9.60 & [$-$10.42, $-$8.85] & no & 12.29 & [11.55, 13.06] \\
\midrule
ResNet 50 & ORT static QDQ & \textbf{4.07} & \textbf{[3.65, 4.53]} & $-$0.15 & [$-$0.51, +0.22] & yes & 3.44 & [3.06, 3.88] \\
 & TRT explicit INT8 & 8.11 & [7.50, 8.73] & $-$1.86 & [$-$2.41, $-$1.34] & no & 6.57 & [6.01, 7.13] \\
 & ET XNNPACK PT2E & \textbf{2.48} & \textbf{[2.16, 2.84]} & $-$0.48 & [$-$0.76, $-$0.19] & yes & 1.73 & [1.48, 2.04] \\
 & LiteRT PT2E static & \textbf{2.42} & \textbf{[2.10, 2.78]} & $-$0.42 & [$-$0.71, $-$0.14] & yes & 1.60 & [1.32, 1.90] \\
\midrule
DeiT tiny & ORT static QDQ & 9.73 & [9.07, 10.38] & $-$0.80 & [$-$1.31, $-$0.32] & no & 11.18 & [10.49, 11.88] \\
 & LiteRT PT2E static & \textbf{3.65} & \textbf{[3.28, 4.07]} & $-$0.06 & [$-$0.38, +0.25] & yes & 4.16 & [3.74, 4.63] \\
\midrule
DistilBERT SST-2 & ORT dynamic & 1.75 & [1.19, 2.37] & $-$0.93 & [$-$1.50, $-$0.36] & no & \multicolumn{2}{c}{n/a} \\
 & ET XNNPACK dynamic & 0.26 & [0.05, 0.52] & +0.05 & [$-$0.15, +0.31] & yes & \multicolumn{2}{c}{n/a} \\
 & LiteRT Quantizer wi8 & 0.67 & [0.31, 1.03] & +0.05 & [$-$0.31, +0.41] & yes & \multicolumn{2}{c}{n/a} \\
 & LiteRT PT2E static & 2.12 & [1.50, 2.79] & $-$0.26 & [$-$0.93, +0.36] & yes & 1.91 & [1.34, 2.53] \\
 & ORT static QDQ & 47.16 & [44.99, 49.43] & $-$38.49 & [$-$41.02, $-$35.96] & no & 26.63 & [24.66, 28.64] \\
\bottomrule
\end{tabular}
\end{table*}

\subsection{The margin locates flips only where accuracy is kept (C2 refuted)}\label{sec:c2}

\begin{figure*}[t]
  \centering
  \includegraphics[width=\textwidth]{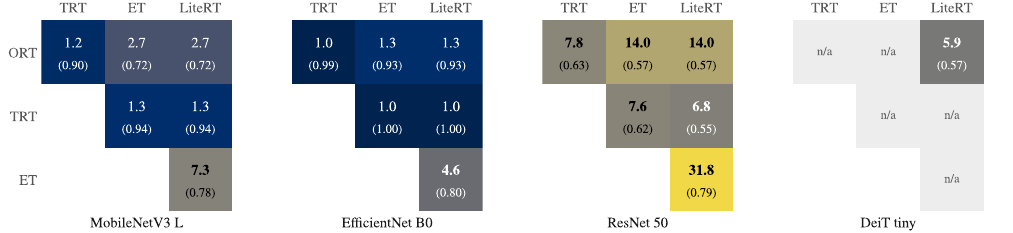}
  \caption{Converters share flipped inputs when INT8 keeps accuracy and stop sharing them when it breaks it.
  Each cell is the lift of two primary INT8 recipes' flip sets on the \nEvalImages{} evaluation images (inputs flipped by
  both over the count expected if flips were independent); in parentheses, the overlap coefficient (shared
  flips over the smaller flip set, exploratory). Bold: 95\% bootstrap lower bound of the lift at least 3, the
  preregistered threshold. n/a: recipe does not build or load (B4, B5). ORT ONNX Runtime, TRT TensorRT,
  ET ExecuTorch.}
  \label{fig:overlap}
\end{figure*}

C2 is refuted as preregistered. Part (a) required an AUROC of at least 0.85 for every vision primary
recipe; it fails on MobileNetV3 (\auMinMobilenetVThreeLarge{} to \auMaxMobilenetVThreeLarge, every 95\% CI inside
[\auLoMobilenetVThreeLarge, \auHiMobilenetVThreeLarge]) and EfficientNet (\auMinEfficientnetBZero{} to \auMaxEfficientnetBZero, inside
[\auLoEfficientnetBZero, \auHiEfficientnetBZero]). Part (b) required a lift of at least 3
for every pair of primary recipes; it fails on the same two models, where pairs reach lifts as low as \liftMinMobilenetVThreeLarge{}
[\liftMinMobilenetVThreeLargelo, \liftMinMobilenetVThreeLargehi] and \liftMinEfficientnetBZero{}
[\liftMinEfficientnetBZerolo, \liftMinEfficientnetBZerohi] and only the ExecuTorch and LiteRT pair, the two recipes with the fewest flips,
passes (\cref{fig:overlap}). Both failures come from recipes that flip a large share of inputs: a recipe that
flips most inputs flips confident ones too, and the lift of two flip sets cannot exceed one over the larger
flip rate, so near 1 is the only possible value.

Where accuracy is kept, both parts hold. On ResNet 50 the margin AUROC is \auMinResnetFiveZero{} to
\auMaxResnetFiveZero{} for every primary recipe (every 95\% CI inside [\auLoResnetFiveZero, \auHiResnetFiveZero]) and
\auMinDeitTiny{} to \auMaxDeitTiny{} on DeiT ([\auLoDeitTiny, \auHiDeitTiny]); on DeiT
\decOneDeitTinyLtPtTwoEIntEightCZero\% [\decOneDeitTinyLtPtTwoEIntEightCZerolo, \decOneDeitTinyLtPtTwoEIntEightCZerohi] of the LiteRT flips fall in the lowest margin decile
(\cref{fig:one}b). Flip sets are shared: ExecuTorch and LiteRT, which both lower PT2E graphs to XNNPACK
style kernels, share ResNet 50 flips \liftResnetFiveZeroEtXnnIntEightCZeroLtPtTwoEIntEightCZero{}
[\liftResnetFiveZeroEtXnnIntEightCZeroLtPtTwoEIntEightCZerolo, \liftResnetFiveZeroEtXnnIntEightCZeroLtPtTwoEIntEightCZerohi] times more
often than independence predicts, and every ResNet 50 pair has a lift of at least \liftMinResnetFiveZero{}
[\liftMinResnetFiveZerolo, \liftMinResnetFiveZerohi].
On text the margin AUROC of the three dynamic recipes is \auDistilbertSstTwoOrtCpuIntEightDynamic{}
[\auDistilbertSstTwoOrtCpuIntEightDynamiclo, \auDistilbertSstTwoOrtCpuIntEightDynamichi] or higher.
We read this as: the margin predicts which inputs a well behaved INT8 recipe flips, and does not
describe a broken one.

\paragraph{Exploratory: no single op type carries most flips.} For ONNX Runtime QDQ we kept one op type at
a time in float, leaving every other scale unchanged, on up to 600 flipped evaluation inputs and as many
unflipped ones per model. The sanity check passes on all four vision models: keeping every quantized op in
float repairs \byRepDeitTiny\% (DeiT), \byRepResnetFiveZero\% (ResNet 50), \byRepMobilenetVThreeLarge\%
(MobileNetV3) and \byRepEfficientnetBZero\% (EfficientNet) of flips and adds at most
\byNewMax\% new ones. No single op type repairs more than \tyMaxAll\% on any model. As a
null we use the share of flips that the calibration redraw alone repairs. On DeiT the best type
(\tyBestDeitTiny, \tyMaxDeitTiny\%) is close to that null (\redrawRepDeitTiny\%), so its flips are spread over
the graph. On the CNNs one type does beat the null: the residual Add on ResNet 50 (\tyMaxResnetFiveZero\%
against \redrawRepResnetFiveZero\%), the Mul nodes of hard swish and squeeze and excitation on MobileNetV3
(\tyMaxMobilenetVThreeLarge\% against \redrawRepMobilenetVThreeLarge\%), and Conv on EfficientNet
(\tyMaxEfficientnetBZero\% against \redrawRepEfficientnetBZero\%); Conv alone repairs about half of the flips on
every CNN (\tyConvResnetFiveZero\%, \tyConvMobilenetVThreeLarge\%, \tyConvEfficientnetBZero\%). Flips are
therefore partly localized on CNNs but never to one op type. Two caveats: keeping one type in float against
scales calibrated for the quantized graph also creates new flips (about half of the unflipped inputs for
EfficientNet Sigmoid), and the MobileNetV3 and EfficientNet INT8 recipes are not accuracy equivalent.

\subsection{A probe set from the source model alone}\label{sec:c3}

\begin{figure}[t]
  \centering
  \includegraphics[width=\columnwidth]{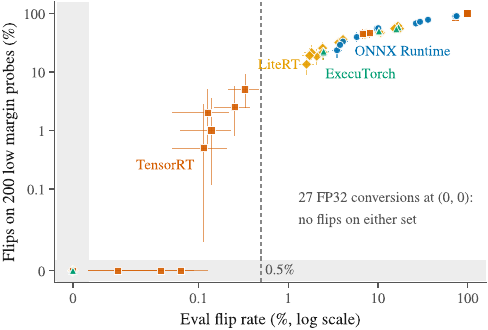}
  \caption{Two hundred low margin probes rank and flag stacks like \nEvalImages{} evaluation images. Points:
  the \nPtsVision{} non control vision stacks; zero has its own shaded row and column. Bars: 95\% intervals (bootstrap over
  images; exact for 200 probes), drawn to the shaded band when they include zero. Every stack right of the dashed 0.5\% line
  flips at least one probe; no FP32 conversion does.}
  \label{fig:probe}
\end{figure}

C3 holds. The probe set is the 200 inputs of the probe pool with the smallest source margin, chosen from
PyTorch outputs before any conversion. Across the \nPtsVision{} non control vision stacks its flip rate
ranks stacks like the full evaluation set (Spearman \rhoPVision; text \rhoPText), and the rule ``flag if
any probe flips'' catches all \nPosVision{} vision and \nPosText{} text stacks with an evaluation flip rate of
at least 0.5\% (recall 95\% CI [\recallVisionlo, \recallVisionhi] and [\recallTextlo, \recallTexthi], exact over stacks;
\cref{fig:probe}). It raises no false alarm on the \nZeroVision{} vision and \nZeroText{}
text FP32 conversions that flip nothing, because a stack that flips no decision on \nEvalImages{} images
also flips none on 200 others. Two hundred random probes do worse: recall \recallRndVision{} [\recallRndVisionlo, \recallRndVisionhi] on vision and
\recallRndText{} [\recallRndTextlo, \recallRndTexthi] on text, Spearman \rhoRndVision{} and \rhoRndText; with
\nPosVision{} and \nPosText{} positive stacks the recall intervals overlap, so the ranking is the clearer difference. The low margin probes are effective
because they concentrate flips: LiteRT INT8 flips \plResnetFiveZeroLtPtTwoEIntEightCZero{} of the 200
ResNet 50 probes against \prResnetFiveZeroLtPtTwoEIntEightCZero{} of 200 random ones.

\subsection{Converter defects met along the way}\label{sec:bugs}

Running only documented recipes we met five defects (\cref{tab:defects}), each logged with a minimal
repro. Two are silent at export and loud at load: ExecuTorch lowers a quantized ViT patch embedding that
XNNPACK then refuses (B4), and the LiteRT dynamic recipe emits an embedding table with a nonzero zero point
that the runtime rejects (B2). A deployment hazard that is not a bug: LiteRT static INT8 returns an int8
output tensor, and a caller that reads it as float without its scale and zero point receives raw codes. None
of the defects changes a reported rate; each removes rows.

\begin{table}[t]
\caption{Five converter defects met while running documented recipes; two pass export and fail only at load.
Minimal repros are in the companion code; none changes a reported rate, they remove rows.}
\label{tab:defects}
\centering\scriptsize\setlength{\tabcolsep}{3pt}
\begin{tabular}{@{}l>{\raggedright\arraybackslash}p{1.9cm}>{\raggedright\arraybackslash}p{2.75cm}>{\raggedright\arraybackslash}p{1.35cm}@{}}
\toprule
 & Converter & Defect & Surfaces \\
\midrule
B1 & ExecuTorch 1.2.0, XNNPACKQuantizer & integer scalars lifted to float tensors, index op fails; fixed on main; patched at run time & loud, calibration \\
B2 & litert\_torch main, dynamic PT2E & int8 embedding with nonzero zero point; runtime rejects it; no LiteRT dynamic row for DistilBERT & silent at export, loud at load \\
B3 & coremltools 9.0 & \texttt{new\_ones}, \texttt{where} not converted; no Core ML DistilBERT; DeiT only via TorchScript & loud, conversion \\
B4 & ExecuTorch 1.2.0, XNNPACK INT8 & ViT patch embedding lowers, XNNPACK refuses a transpose; no ExecuTorch INT8 row for DeiT & silent at export, loud at load \\
B5 & TensorRT 10.16 ONNX parser & per tensor DequantizeLinear on a 0-d scale rejected (axis check); no TensorRT INT8 for DeiT, DistilBERT & loud, parse \\
\bottomrule
\end{tabular}
\end{table}

\section{Related work}\label{sec:related}

The closest work holds one INT8 artifact fixed and varies the hardware.
\citet{shin2026int8portable} runs one ONNX QDQ model with fixed scales on seven hardware classes and
finds FP32 top 1 identical everywhere and INT8 decisions identical only when two targets share an integer
kernel, with accuracy preserved. We hold the hardware class fixed and vary the converter and its own
documented recipe, which is the choice a deployment team makes, compare against the PyTorch source rather
than between targets, add a vision transformer and a text model with intervals and a calibration redraw
floor, and test a probe set. \citet{chen2026integeralibi} swap only the INT8 GEMM kernel inside an LLM
serving engine, find that flips concentrate at small logit margins, and propose kernel conformance tests;
\citet{chen2026pow2} add power of two scales and tolerance based conformance. Our C2 result tests their
margin mechanism for whole model converters and finds it holds only where accuracy is kept.

Software engineering work treats converter disagreement as a fault. \citet{louloudakis2023faultloc}
localize and repair label differences between source and converted image models;
\citet{jajal2024onnx} survey ONNX converter failures and find a third are semantically incorrect models;
\citet{openja2022converting} report that ONNX and Core ML conversions keep accuracy; \citet{chen2021mobiledeploy}
build a taxonomy of mobile deployment faults; DiffChaser \citep{xie2019diffchaser} searches for synthetic
inputs on which converted or quantized models disagree; \citet{li2025crossbackend} verify cross backend
consistency at tensor and task level. We measure, on natural held out inputs and with intervals, the
disagreement that remains when every converter works as documented, and how well a cheap probe set detects
it.

\section{Limitations}\label{sec:limits}

\begin{enumerate}
\item \textbf{CPU proxy for phone kernels.} ExecuTorch and LiteRT ran their x86 kernels; ARM kernels may
round differently. Direction unknown; a phone run would remove it.
\item \textbf{Core ML not executed.} Core ML is conversion only; its flip rate is unmeasured.
\item \textbf{One recipe per converter is ours to choose.} We used each converter's documented default;
tuned recipes (per layer exclusion, more calibration data) would likely flip fewer inputs.
\item \textbf{Calibration noise is as large as the effect.} With two draws we can bound but not average it;
flip rates of single builds may be biased in either direction by up to the K4 rate.
\item \textbf{Data.} Twenty ImageNet classes and a binary sentiment task; per class difficulty differs on
the full 1000 classes, likely raising flip rates there.
\item \textbf{Missing rows.} Defects B4 and B5 remove three primary recipes on DeiT and TensorRT INT8 on
DistilBERT; ExecuTorch portable kernels did not run on ResNet 50.
\end{enumerate}
We do not claim that any converter is wrong or that its INT8 models are inaccurate; we claim that accuracy
does not certify a conversion and that a 200 input probe set chosen from the source model does.

\section*{Impact Statement}
On device models make decisions for individual users, and a conversion that keeps accuracy can still flip
those decisions for a few percent of them without anyone noticing. The probe set we propose is cheap enough
to run in a release pipeline and makes such flips visible before they ship. We see no negative use
specific to this work.

\section*{AI Disclosure Statement}
An AI assistant was used to write and debug experiment code, to draft and edit the text, and to search for
related work. Every number in the paper is generated by scripts from saved outputs; every reference was
opened at its source and checked by the author.

\bibliography{refs}
\bibliographystyle{icml2027}
\end{document}